\newif\ifacm
\acmfalse 
\newcommand{\acmOrArxiv}[2]{\ifacm #1\else #2\fi}

\acmOrArxiv{%
  \documentclass[sigconf]{acmart}
}{%
  \documentclass[sigconf, authorversion]{acmart}
}

\copyrightyear{2022}
\acmYear{2022}
\setcopyright{acmlicensed}
\acmConference[KDD '22]{Proceedings of the 28th
ACM SIGKDD Conference on Knowledge Discovery and Data Mining}{August
14--18, 2022}{Washington, DC, USA}
\acmBooktitle{Proceedings of the 28th ACM SIGKDD Conference on Knowledge
Discovery and Data Mining (KDD '22), August 14--18, 2022, Washington, DC,
USA}
\acmPrice{15.00}
\acmDOI{10.1145/3534678.3539165}
\acmISBN{978-1-4503-9385-0/22/08}

\acmOrArxiv{\settopmatter{authorsperrow=5}}{}

\AtBeginDocument{%
  }

\begin{document}

\title{Greykite: Deploying Flexible Forecasting at Scale at LinkedIn}


\acmOrArxiv{%
  \author{Reza Hosseini}
  \affiliation{}
  \authornote{Both authors contributed equally to this research.}
  \author{Albert Chen}
  \affiliation{}
  \authornotemark[1]
  \author{Kaixu Yang}
  \affiliation{}
  \author{Sayan Patra}
  \affiliation{}
  \author{Yi Su}
  \affiliation{}
  \author{Saad Eddin Al Orjany}
  \affiliation{}
  \author{Sishi Tang}
  \affiliation{%
    \institution{LinkedIn Corp}
    \streetaddress{605 W Maude Ave}
    \city{Sunnyvale}
    \state{CA}
    \country{USA}
    \postcode{94085}
  }
  \author{Parvez Ahammad}
  \affiliation{}
}{%
  \author{Reza Hosseini$^*$, Albert Chen$^*$, Kaixu Yang, Sayan Patra, Yi Su\\ Saad Eddin Al Orjany, Sishi Tang, Parvez Ahammad}
  \affiliation{%
    \institution{LinkedIn Corporation}
    \city{Sunnyvale}
    \state{CA}
    \country{USA}
  }
  \email{{rhosseini, abchen, kayang, sapatra, ysu1, sorjany, sistang, pahammad}@linkedin.com}
}

\acmOrArxiv{%
  \renewcommand{\shortauthors}{Reza Hosseini et al.}
}{%
  \renewcommand{\shortauthors}{Hosseini and Chen et al.}
}

\begin{abstract}
Forecasts help businesses allocate resources and achieve objectives.
At LinkedIn, product owners use forecasts to set business targets, track outlook, and monitor health. Engineers use forecasts to efficiently provision hardware.
Developing a forecasting solution to meet these needs
requires accurate and interpretable forecasts on diverse time series with sub-hourly to quarterly frequencies.
We present \texttt{Greykite}, an open-source Python library for forecasting
that has been deployed on over twenty use cases at LinkedIn.
Its flagship algorithm, Silverkite, provides interpretable, fast, and highly flexible univariate forecasts
that capture effects such as time-varying growth and seasonality, autocorrelation, holidays, and regressors.
The library enables self-serve accuracy and trust by facilitating data exploration, model configuration, execution, and interpretation.
Our benchmark results show excellent out-of-the-box speed and accuracy on datasets from a variety of domains.
Over the past two years, \texttt{Greykite} forecasts have been trusted by Finance, Engineering, and Product teams for
resource planning and allocation,
target setting and progress tracking,
anomaly detection and root cause analysis.
We expect \texttt{Greykite} to be useful to forecast practitioners with similar applications who need accurate, interpretable forecasts that capture complex dynamics common to time series related to human activity.
\end{abstract}



\maketitle
    \acmOrArxiv{}{%
      \thispagestyle{firstpage}
      \def\thefootnote{*}\footnotetext{Both authors contributed equally to this research.}\def\thefootnote{\arabic{footnote}}
    }
    \emergencystretch 3em

    \section{Introduction}
\label{sect:introduction}
Time series forecasts aim to provide accurate future expectations
for metrics and  other quantities that are measurable over time. 
At LinkedIn, forecasts are used for performance management and resource management.
Performance management is the process of setting business metric targets and
tracking progress to ensure we achieve them. This is the heartbeat of how LinkedIn
manages its business \cite{nextplay}.
Resource management involves optimizing allocation of budget, hardware, and other resources,
often based on forecasts of business and infrastructure metrics.

Before this work, the processes for resource and performance management
at LinkedIn were highly manual, relying on rule-based spreadsheets,
simple linear regressions, or expert opinions.
Domain knowledge and expert judgment, while often accurate,
are subjective, hard to validate over time, and very hard to scale.
Furthermore, they have trouble adapting to ecosystem shocks (such as COVID-19) that
change underlying dynamics and invalidate existing approaches to estimate
metric growth. We have developed a forecasting library and framework that bring
accuracy, scale, and consistency to the process. Furthermore, algorithmic forecasts
can be automated and therefore integrated into systems that derive additional
insights for decision-making.

\section{Motivation and Related Work}
\label{sect:motivation}
LinkedIn has diverse forecast requirements, with 
data frequencies from sub-hourly to quarterly (fiscal quarter) and
forecast horizons from
15 minutes
to over a year.
Many of our time series exhibit strong growth and seasonality patterns
that change over time due to product launches, ecosystem dynamics, or long-term drifts.
Seasonality here 
refers to periodic/cyclic patterns over multiple horizons, e.g.\;daily seasonality refers
to a cyclic pattern over the course of a day.
For example, a new recommendation system may change the delivery schedule of mobile 
app notifications, affecting the seasonality pattern of user engagement;
external shocks like COVID-19 can have an effect on hiring, job-seeking activity,
and advertiser demand; long-term changes may shift engagement from desktop to mobile
and affect traffic to our services.
New features are constantly impacting the metrics, as
LinkedIn runs hundreds of concurrent A/B tests each day.

A suitable forecast algorithm must account for time series
characteristics such as:
(a) strong seasonality, with periods from daily to yearly,
(b) growth and seasonality changes,
(c) high variability around holidays, month/quarter boundaries,
(d) local anomalies from external events and engineering bugs,
(e) floating holidays (those with irregular intervals),
(f) dependency on external factors such as macroeconomic indicators.
It must also provide reliable prediction intervals to quantify uncertainty
for decision making and anomaly detection.
We expect such requirements to be common across many industries, and certainly
within the technology sector, where time series patterns depend on human activity
(of users) and product launches, and where forecasts are needed for both long-term
planning and short-term monitoring.

Our goal is to deliver a self-serve forecasting solution to data scientists
and engineers across the company, many of whom have no specialized forecasting expertise.
Developers need a way to develop accurate forecasts with little effort. Stakeholders need
to trust the forecasts, which requires answering questions such as: 
Does the forecast include impact from the latest product launch?
How quickly does the forecast adapt to new patterns?
What is the expected impact of the upcoming holiday?
Why has the forecast changed?
Can the forecast account for macroeconomic effects?
What would happen if economic recovery is fast/slow?
To answer these questions, the model must be interpretable and easily
tuned according to expert input when available.

The area of forecasting time series has a long history,
with many models and techniques developed in the past decades.
Some important examples include:
classical time series models such as ARIMA
(e.g.\;\cite{hyndman2018forecasting}) and GARCH \cite{book-tsay-2010};
exponential smoothing methods \cite{winters-1960};
state-space models \cite{kalman-1960, book-durbin-2012, west2006bayesian};
generalized linear model extensions to time series
\cite{book-kedem-2002, hosseini-takemura-2015};
deep learning models such as RNN \cite{salinas2020deepar}.
There are several popular open-source forecasting packages.
Notable examples include
\texttt{fable} \cite{fable},
\texttt{statsmodels} \cite{seabold2010statsmodels},
\texttt{Prophet} \cite{taylor-prophet-2018},
and \texttt{GluonTS} \cite{gluonts_jmlr}.
The list of packages has grown quickly in recent years due to 
high demand for reliable forecasting in many domains.
Each package has a different focus within the forecasting landscape.

\texttt{fable} and \texttt{statsmodels} implement statistical models such as ARIMA and ETS,
which can be trained and scored quickly.
The ARIMA model captures temporal dependence and non-stationarity, and has
extensions to include seasonality and regressors. However, it is not flexible enough
to capture effects such as seasonality that changes in magnitude and shape over time,
or volatility that increases around certain periods such as holidays or
weekends. Nor is it readily interpretable for business stakeholders.
ETS, while more interpretable than ARIMA, has similar problems capturing
interaction effects and does not support regressors.

\texttt{Prophet} is a popular forecasting library that includes
the Prophet model, designed to capture trend, seasonality, trend changepoints,
and holidays. The package allows visualization of these effects for interpretability.
However, model training and inference are slower due to its Bayesian formulation.
And while the interface is user-friendly, with only a few tuning parameters,
the model is not flexible enough to achieve high accuracy on complex time series.
For example, it does not perform as well for short-term forecasts due to lack of
native support for autoregression or other mechanisms to capture short-term dynamics.
Prophet supports custom regressors, but these must be provided by the user
for both training and inference; this is inconvenient for standard time features
and cumbersome for complex interaction terms.

\texttt{GluonTS} is a time series modeling library that includes the
DeepAR algorithm \cite{salinas2020deepar}. DeepAR is a deep
learning model that trains a single global model on multiple time series to forecast.
While this can be powerful for automated forecasting, the cross-time
series dependencies between input and output and lack of intuitive parameters
make it hard to interpret the model or apply expert knowledge to improve the forecast.

To support LinkedIn’s forecasting needs, we developed \texttt{Greykite},
a Python library for self-serve forecasting.\footnote{\url{https://github.com/linkedin/greykite}}
Its flagship algorithm, Silverkite, provides univariate
forecasts that capture diverse time series characteristics with speed and interpretabilty.
Our contributions include:
(1) flexible design to capture complex time series dynamics for any frequency and forecast horizon,
(2) forecast components that can be explicitly tuned,
(3) interpretable output to explain forecast drivers,
(4) fast training and inference,
(5) decoupled volatility model that allows time-varying prediction intervals,
(6) flexible objective function to predict peak as well as mean,
(7) self-serve forecasting library that facilitates data exploration, model configuration, tuning, and diagnostics.

    \section{Methodology}
\label{sect:methodology}
\begin{figure*}[h]
    \vspace{-0.1in}
    \centering
    \includegraphics[width=\linewidth]{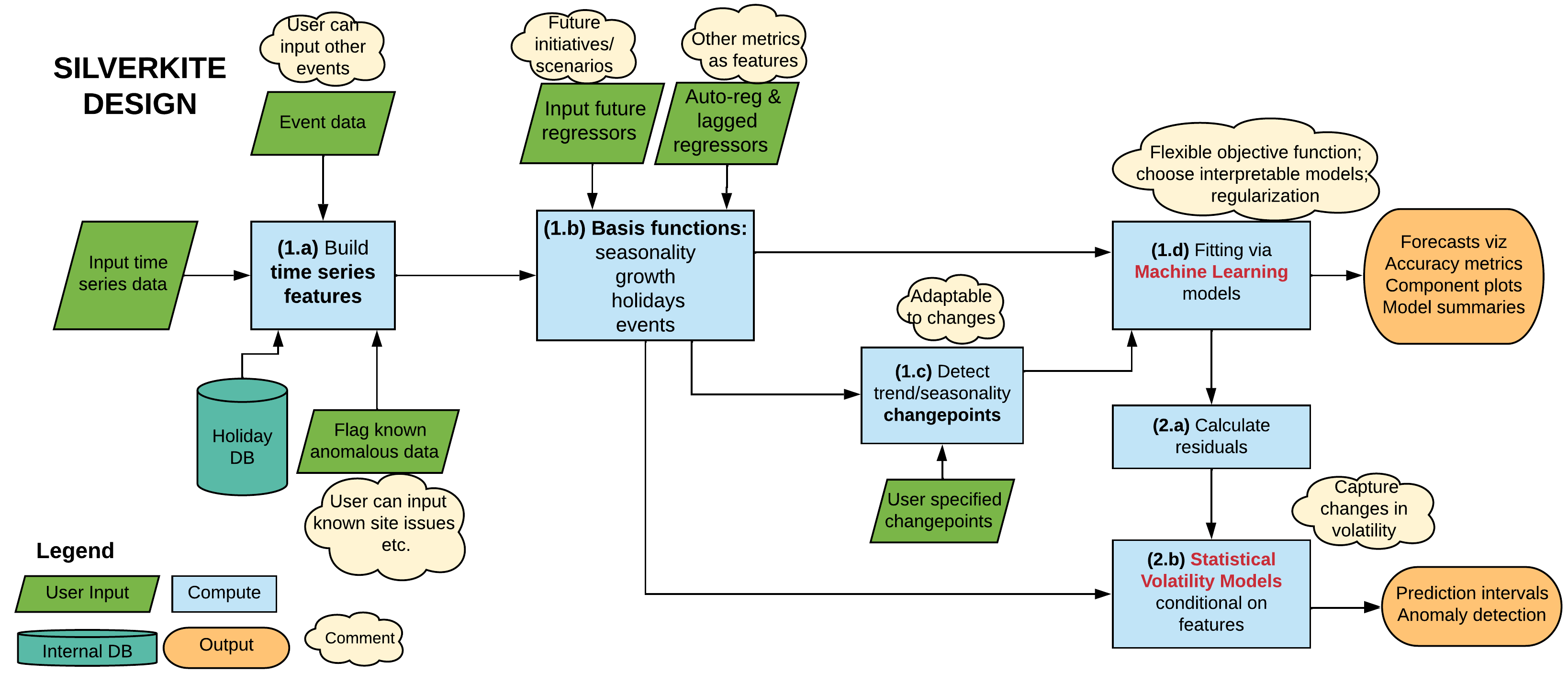}
    \caption{Architecture diagram for \texttt{Greykite}'s main forecasting algorithm: Silverkite.}
    \label{fig:architecture}
    \Description{}
    \vspace{-0.1in}
\end{figure*}

In this section, we explain the design of Silverkite, the
core forecasting algorithm in the \texttt{Greykite} library. The Silverkite algorithm 
architecture is shown in Figure~\ref{fig:architecture}.
The user provides the input time series and any known anomalies, events, 
regressors, or changepoint dates. The model returns forecasts, prediction
intervals, and diagnostics.
The computation steps of the algorithm are decomposed into two phases:
\vspace{-0.03in}
\begin{itemize}
    \item Phase (1): the conditional mean model.
    \item Phase (2): the volatility/uncertainty model.
\end{itemize}
In (1), a model predicts the metric of interest, and in (2), a volatility model is fit to the residuals. This design helps us with flexibility and speed, because integrated models are often more susceptible to poor tractability (convergence issues for parameter estimates) or divergence in the predicted values.
Phase (1) can be broken down into these steps:
\vspace{-0.03in}
\begin{itemize}
    \item[(1.a)] Extract raw features from timestamps, events data, and history of the series (e.g. hour, day of week).
    \item[(1.b)] Transform the features to appropriate basis functions (e.g. Fourier series terms for various time scales).
    \item[(1.c)] Apply a changepoint detection algorithm to the data to discover changes in the trend and seasonality over time.
    \item[(1.d)] Apply an appropriate machine learning algorithm to fit the covariates from (1.b) and (1.c) (depending on the objective).
\end{itemize}
Step (1.b) transforms the features into a space which can be used in additive models
for better interpretability. For Step (1.d), we recommend regularization-based models such as ridge regression for mean prediction
and (regularized) quantile regression for quantile prediction (e.g.\;for peak demand).
In Phase (2), a conditional variance model is fit to the residuals, which allows the volatility to be
a function of specified categories, e.g. day of week.

\subsection{The Model Formulation}
\label{subsection:model_formulation}
In this subsection, we give a brief overview of the mathematical formulation of the Silverkite model.
We will demonstrate how this works on a real example in Section~\ref{sect:modeling_framework}.

Suppose $\{Y(t)\}, t=0,1,\ldots$ is a real-valued time series where $t$ denotes time.
Following \cite{book-kedem-2002} and \cite{hosseini-2011-categ}, we denote the available information up to time $t$ by $\mathcal{F}(t)$. For example $\mathcal{F}(t)$ can include covariates such as $Y(t-1), Y(t-2), X(t), X(t-1)$ where $Y(t-i)$ denotes lags of $Y$; $X(t)$ is the value of a regressor known at time $t$ and $X(t-1)$ is the value of the same regressor at time $t-1$. The latter is often referred to as lagged regressor.

\subsubsection*{The conditional mean model}
The conditional mean model is
\begin{align}
    \mathbb{E}[Y(t) | \mathcal{F}(t)] \sim G(t) + S(t) + H(t) + A(t) + R(t) + I(t),
\label{eq:mean-model}
\end{align}
where $G, S, H, A, R, I$ are functions of covariates in $\mathcal{F}(t)$.
They are linear combinations of these covariates or their interactions.

$G(t)$ is the general growth term that may include trend changepoints $t_1,\ldots,t_k$, as
\vspace{-0.05in}
\[
    G(t)= \alpha_0 f(t) + \sum_{i=1}^k \alpha_i \mathbf{1}_{\{t > t_i\}} (f(t) - f(t_i)),
\]
where $f(t)$ is any growth function, e.g. $f(t)=t$ or $f(t)=\sqrt{t}$ and $\alpha_i$'s are parameters to be estimated. 
Note that $G(t)$ is a continuous and piecewise smooth function of $t$.

$S(t)=\sum_{p\in\mathcal{P}}S_p(t)$ includes all Fourier series bases for different seasonality components (yearly seasonality, weekly seasonality, etc.), where $\mathcal{P}$ is the set of all seasonal periods. A single seasonality component $S_p(t)$ can be written as
\vspace{-0.05in}
\[
    S_p(t) = \sum_{m=1}^M [a_m\sin (2m\pi d(t)) + b_m\cos (2m\pi d(t))],
\]
where $a_m,\;b_m$ are Fourier series coefficients to be estimated by the model and $M$ is the series order. $d(t)\in[0, 1]$ is the corresponding time $t$ within a seasonal period. For example, daily seasonality has $d(t)$ equal to the time of day (in hours divided by 24) at time $t$. Moreover, for a list of time points $t_1,\ldots,t_K$ where the seasonality is expected to change in either shape or magnitude or both, Silverkite models these changes as
\vspace{-0.05in}
\[
    S_{cp}(t) = S_p(t; \{a_m, b_m\}) + \sum_{k=1}^K\mathbf{1}_{\{t>t_k\}} S_p(t; \{a_{mk}, b_{mk}\}),
\]
where $S_{cp}(t)$ is a single seasonality component and $S_p$ is the seasonality term with coefficients $\{a_{mk}, b_{mk}\}$.
Similar to trend changepoints, this formulation allows the Fourier series'
coefficients to change and adapt to the most recent seasonal patterns.
$S_p(t)$ can also be modeled with categorical variables such as hour of day.

$H(t)$ includes indicators on holidays/events and their neighboring days.
For example, $\mathbf{1}_{\{t\in\text{Christmas day}\}}$ models Christmas effect.
Holidays can have extended impact over several days in their proximity.
Silverkite allows the user to customize the number of days before and after the event where the
impact is non-negligible, and models them with separate indicators/effects.
Indicators on month/quarter/year boundaries also belong to this category.

$A(t)$ includes any time series information known up to time $t$ to model the remaining time dependence. For example, it can be lagged observations $Y(t-1), \ldots, Y(t-r)$ for some order $r$, or the aggregation of lagged observations such as $AVG(Y(t); 1,2,3) = \Sigma_{i=1}^3 Y(t-i) / 3$. Aggregation allows for parsimonious models that capture long-range time dependencies.

$R(t)$ includes other time series that have the same frequency as $Y(t)$ that can meaningfully be used
to predict $Y(t)$. These time series are regressors, denoted $X(t) = \left\{X_1(t), \ldots , X_p(t)\right\},\ t=0, 1, \ldots$
in the case of $p$ regressors.
If available forecasts $\hat{X}(t)$ of $X(t)$ are available, let $R(t) = \hat{X}(t)$.
Otherwise, lagged regressors or aggregations of lagged regressors can be used,
such as $R(t) = X(t - h)$ where $h$ is the forecast horizon. If the minimum lag order is at least the forecast horizon, the inputs needed for forecasting have already been observed. Otherwise, in autoregression, having lag orders smaller than forecast horizon means that forecasted values must be used. To handle this, we incrementally simulate the future values needed for calculating the later forecasted values.

$I(t)$ includes any interaction of the above terms to model complex patterns. For example, we can model 
different daily seasonality patterns during weekend and weekdays by including
\[
    \mathbf{1}_{\{t\in\text{weekend}\}} \times \text{daily seasonality Fourier series terms},
\]
where $\times$ denotes interaction between two components.

To mitigate the risk of the model becoming degenerate or over-fitting, regularization can be 
used in the machine learning fitting algorithm for the conditional model. In fact, regularization 
also helps in minimizing the risk of divergence of the simulated future series for the model \cite{hosseini-bk-2020, hosseini-takemura-2015, book-tong-1990}.

\subsubsection*{Automatic changepoint detection}
\label{sussubsect:changepoint}
The trend and seasonality changepoints help Silverkite stay flexible and adapt to the most recent growth and seasonal patterns. Silverkite offers fast automatic changepoint detection algorithms to reduce manual modeling efforts.
Automatic changepoint detection is described below.

For trend changepoints, we first aggregate the time series into a coarser time series to eliminate short-term
fluctuations (which are captured by other features, such as holidays). For example, daily data can be aggregated
into weekly data. Next, a large number of potential trend changepoints are placed evenly across time, except for a
time window at the end of the time series to prevent
extrapolation from limited data.
We model the aggregated time series as a function of trend, while controlling for yearly seasonality:
\[
    Y_a(t) \sim G(t) + S_y(t),
\]
where $Y_a(t)$ is the aggregated time series, $G(t)$ is the growth with all potential trend changepoints and $S_y(t)$ is yearly seasonality.
The adaptive lasso penalty is used to identify significant trend changepoints \cite{zou2006adaptive}
(lasso would over-shrink significant changepoints’ coefficients to 
reach the desired sparsity level \cite{tibshirani1996regression, zou2006adaptive}).
Finally, identified changepoints that are too close to each other are merged to improved model stability,
and the detected changepoints are used to construct the piecewise growth basis function.
See \cite{hosseini-greykite-2021} for implementation details.

For seasonality changepoints, we use a similar formulation:
\[
    Y_{d}(t) \sim \sum_{p\in\mathcal{P}}S_{cp}(t),
\]
where $Y_d(t)$ is the de-trended time series and $S_{cp}(t)$ is defined in Section~\ref{subsection:model_formulation}
, with $t_1, \ldots, t_K$ being potential seasonality changepoints.
Automatic selection is also done with adaptive lasso.
The formulation allows seasonality estimates to change in both pattern and magnitude.
This approach can capture similar effects as multiplicative seasonality, but
is far more flexible in terms of the pattern changes,
and avoids the problems of multiplicative seasonality magnitude
growing too large with longer forecast horizons.

\subsubsection*{The volatility model}
The volatility model is fit separately from the conditional mean model, 
with the following benefits compared to an integrated model
that estimates mean and volatility jointly:
\begin{enumerate}
    \item[(a)] Stable parameter estimates and forecasts (e.g.\;\cite{book-tong-1990, hosseini-bk-2020}).
    \item[(b)] Speed gain by avoiding computationally heavy Maximum Likelihood Estimation
    (e.g.\;\cite{hosseini-takemura-2015}) or Monte Carlo methods (e.g.\;\cite{west2006bayesian}).
    \item[(c)] Flexible to pair any conditional mean model with any volatility model for better accuracy.
    \item[(d)] Modular engineering framework for simple development and testing.
\end{enumerate}

These factors are important in a production environment
that requires speed, accuracy, reliability, and code maintainability.
It is typical to see larger volatility around certain periods such as holidays and month/quarter boundaries.
Silverkite's default volatility model uses conditional prediction intervals to adapt to such effects,
allowing volatility to depend on categorical features. The model is described below.

Let $Y(t)$ be the target series and $\hat{Y}(t)$ be the forecasted series. Define the residual series
as $r(t) = Y(t) - \hat{Y}(t)$. Assume the volatility depends on given categorical features $F_1, \ldots, F_p$
which are also known into the future, for example, day of week. Consider the empirical distribution $(R | F_1, \ldots, F_p)$
and fit a parametric or non-parametric distribution to the combination as long as the sample size for that combination, denoted
by $n(F_1, \ldots, F_p)$ is sufficiently large e.g.\; $n(F_1, \ldots, F_p) > N,\; N=20$.
Note that one can find an appropriate $N$ using data (for example, during cross-validation by checking the
distribution of the residuals). Then from this distribution, we estimate the quantiles $Q(F_1, \ldots, F_p)$
to form the prediction interval with level $1-\alpha$:
\vspace{-0.05in}
\[
    (\hat{Y}(t) + Q(F_1, \ldots, F_p)(\alpha/2),\;\hat{Y}(t) + Q(F_1, \ldots, F_p)(1-\alpha/2)).
\]
One choice for a parametric distribution is the Gaussian distribution 
$\mathcal{N}(0, \sigma^2(F_1, \ldots, F_p))$. While the residuals of a naive model can be far from normal, it is possible that
after conditioning on the appropriate features, the residuals of a sufficiently complex mean model are approximately normal as observed in \cite{hosseini-takemura-2015}. This assumption can be checked by inspecting the qq-plots of the conditional errors. Silverkite offers an option to use empirical quantiles to construct the prediction intervals when this assumption is violated.
Silverkite's flexibility allows other volatility models to be added.
For example, a regression-based volatility model could be used to condition on many features, including continuous ones.

\subsection{How Silverkite Meets the Requirements}
Silverkite handles the time series characteristics mentioned in Section~\ref{sect:motivation}.
Strong seasonality is captured by the Fourier series basis function $S(t)$.
A higher order $M$ or categorical covariates (e.g. hour of day) can be used to
capture complex seasonality patterns.
Growth and seasonality changes over time are handled by automatic detection of
trend and seasonality changepoints. Additionally, autoregression allows quick
adaptation to new patterns, which is especially useful in short-term forecasts.
High variability around holidays and month/quarter boundaries is addressed by explicitly
including their effects in the mean model, and by allowing the volatility model to
condition on such features.
To capture changes in seasonality on holidays, Silverkite allows for interactions between
holiday indicators and seasonality.
Floating holidays are easily handled by looking up their dates in \texttt{Greykite}'s holiday database.
Local anomalies are handled by removing known anomalies from the training set.
Impact of external factors is incorporated through regressors, whose forecasted values
can come from Silverkite or another model. This allows for comparing forecast scenarios.
Thus, Silverkite's design captures these time series characteristics in an intuitive manner amenable to modeling and interpretation.

    \section{Modeling framework}
\label{sect:modeling_framework}
\begin{figure}[h]
    \vspace{-0.05in}
	\centering
	\includegraphics[width=\linewidth]{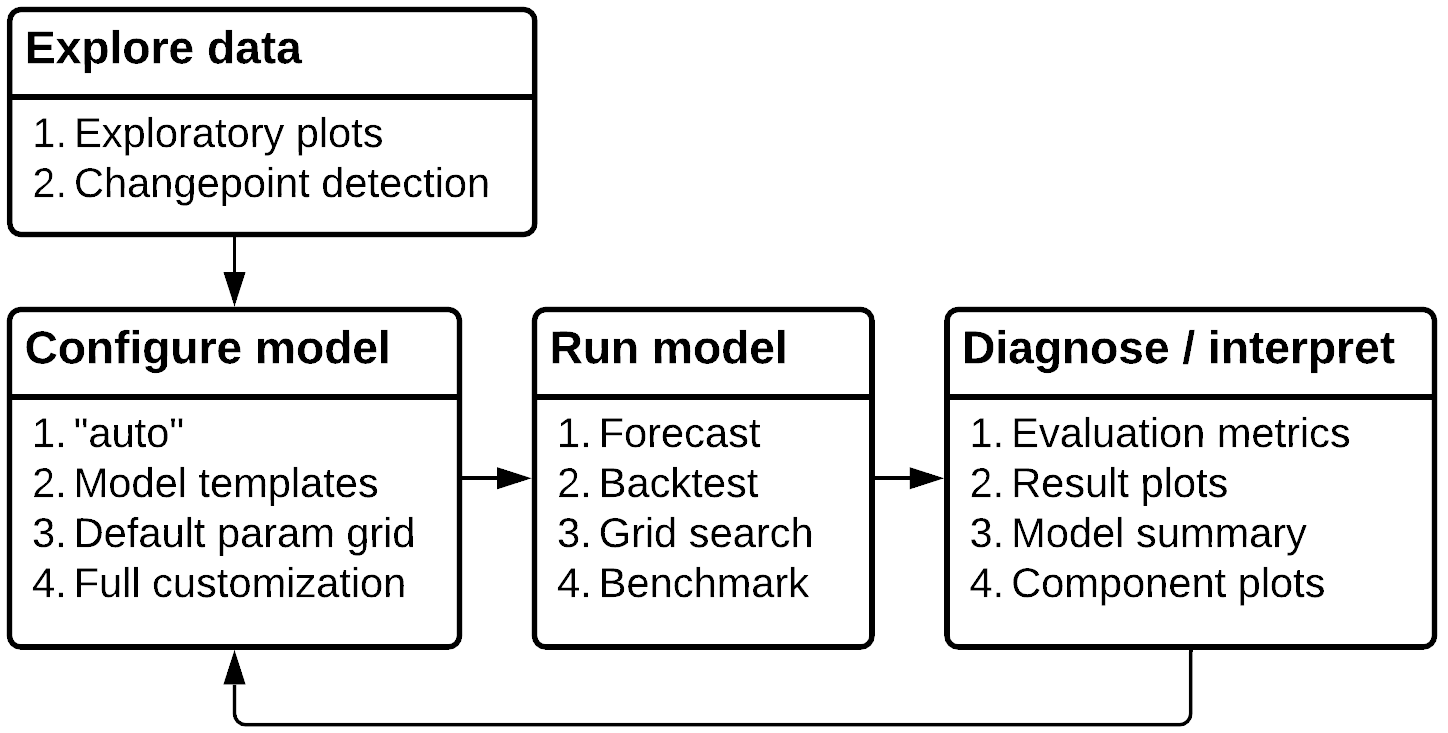}
	\caption{Library aids each step of the forecast workflow.}
	\label{fig:model_workflow}
	\Description{}
	\vspace{-0.1in}
\end{figure}

We implemented these algorithms in \texttt{Greykite}, a Python library
for data scientists and engineers across the company.
Figure ~\ref{fig:model_workflow} shows how the library facilitates
each step of the modeling process.

\begin{figure}[h]
    \vspace{-0.1in}
    \centering
    \includegraphics[width=\linewidth]{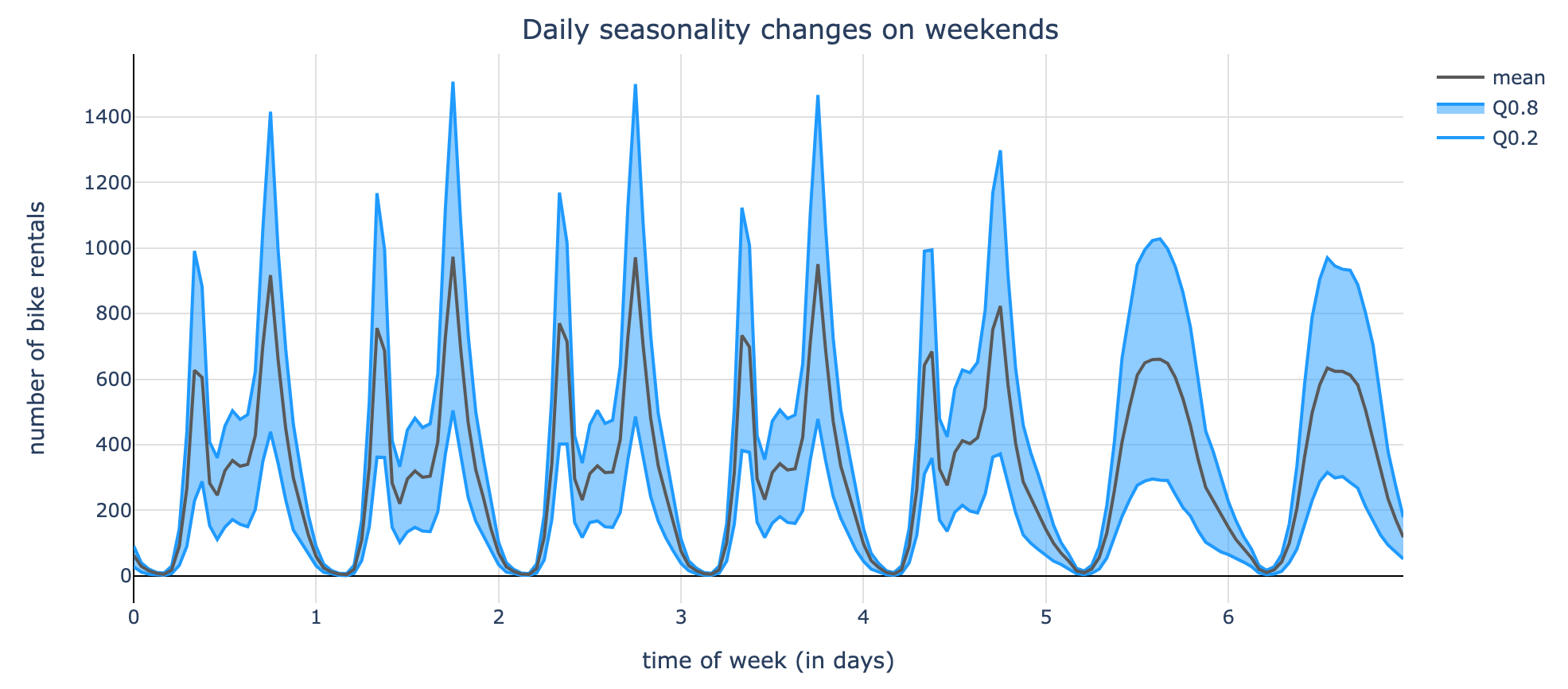}
    \includegraphics[width=\linewidth]{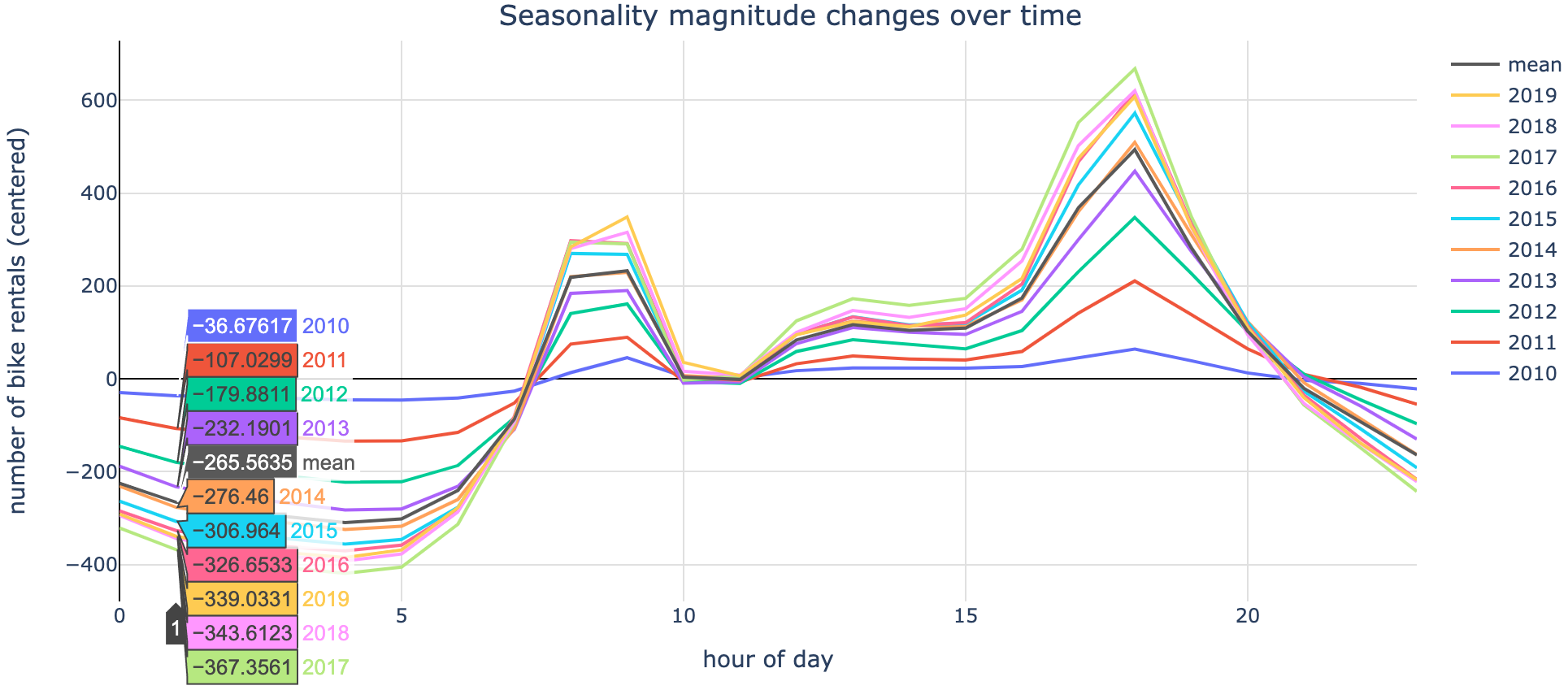}
    \caption{\texttt{Greykite} provides exploratory plots to identify seasonality effects and interactions.
    }
    \label{fig:bikesharing_fine_tuning}
    \Description{}
    \vspace{-0.1in}
\end{figure}

The first step is data exploration.
\texttt{Greykite} provides interactive, exploratory plots to assess seasonality, trend, and holiday effects.
These plots help users to identify effects that are hard to see when plotting the entire time series.
We illustrate this with the D.C. bike-sharing time series \cite{bikeshare} in Figure~\ref{fig:bikesharing_fine_tuning}.
The top plot shows the daily seasonality pattern across the week.
The shape changes between weekends and weekdays, revealing an
interaction between daily seasonality and \texttt{is\_weekend}.
The bottom plot shows the (mean centered) yearly seasonality and an overlay for each year.
By comparing how the lines change over time, we can see that seasonality magnitude
increased until 2017, then decreased.

\begin{figure}[h]
    \centering
    \includegraphics[width=\linewidth]{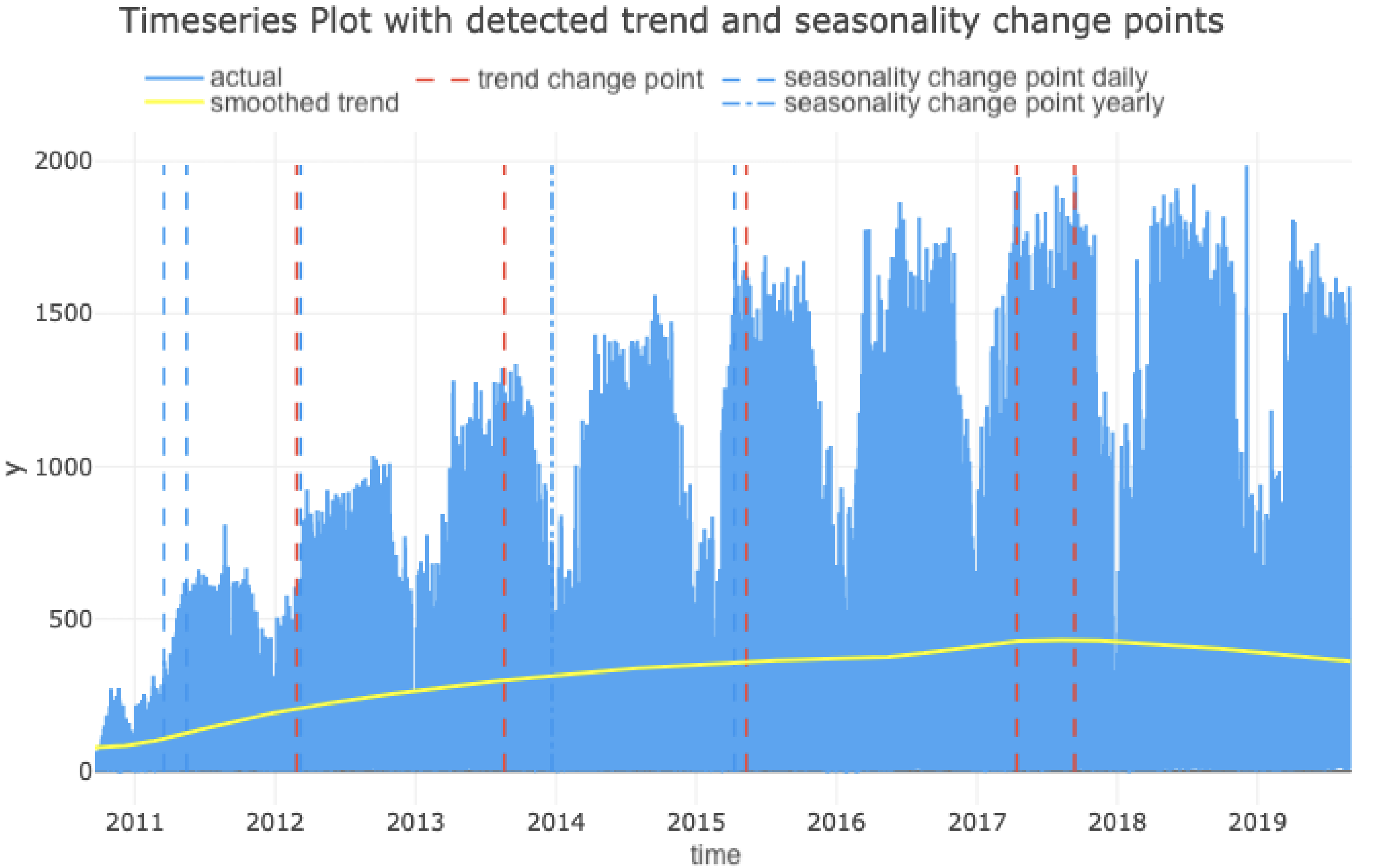}
    \caption{Automatic trend and seasonality changepoint detection in bike-sharing data.}
    \label{fig:bikesharing_changepoints}
    \vspace{-0.2in}
\end{figure}

The automatic changepoint detection module can be used on its own to explore trend and seasonality.
Figure~\ref{fig:bikesharing_changepoints} shows a changepoint detection result,
revealing both trend and seasonality changepoints in the time series.

The second step is configuration.
The user provides time series data and a forecast config
to the $\texttt{Forecaster}$ class, which produces the
forecast, diagnostics, and the trained model object.
The config allows the user to select an algorithm,
forecast horizon, coverage, model tuning parameters,
and evaluation parameters. The config is optional to make quick
start easy, but also very customizable.

\texttt{Greykite} exposes different algorithms, including Silverkite,
Prophet,
and SARIMA,
as \texttt{scikit-learn} estimators that can be configured from the forecast config.
Silverkite provides intuitive tuning parameters such as: autoregressive lags,
changepoint regularization strength, and the list of holiday countries. Many Silverkite tuning parameters,
including changepoints, seasonality, autoregression, and interaction terms
have an intelligent ``auto'' setting. Others have reasonable defaults,
such as ridge regression for the machine learning model.

For high-level tuning, we introduce the concept of \emph{model templates}.
Model templates define forecast parameters for various data characteristics and
forecast requirements (e.g. hourly short-term forecast, daily long-term forecast).
Model templates drastically reduce the search space to find a satisfactory forecast.
They allow decent forecasts out-of-the-box even without data exploration.
When model template is ``AUTO'', \texttt{Greykite} automatically selects the best model template for the input data.

Fine-tuning is important to get the best possible accuracy for key business metrics with
high visibility and strict accuracy requirements. Therefore, our library 
provides full flexibility to customize the settings of a model template.
For example, the user can add custom changepoints to enable faster adaptation to known changes
and label known anomalies in the training data.
The user can easily experiment with derived features by specifying model formula terms such
as \texttt{`is\_weekend:y\_lag\_1'} (weekend/AR1 interaction). Because Silverkite generates
many features internally, the user can leverage these to fine-tune the model without writing any code.

The third step is running the model.
Internally, the $\texttt{Forecaster}$ class runs an end-to-end forecasting pipeline
with pre-processing, hyperparameter grid search, evaluation, backtest, and forecast.
Grid search enables forecasting many metrics in a more automated way
by selecting the optimal model from multiple candidates.
We offer a default parameter grid for efficient search of the space.
Automated machine learning techniques may also be used.
Silverkite's fast training and inference facilitates such hyperparameter tuning.
\texttt{Greykite} offers a benchmarking class to compare algorithms.

\begin{figure}[h]
    \vspace{-0.1in}
	\centering
	\includegraphics[width=\linewidth]{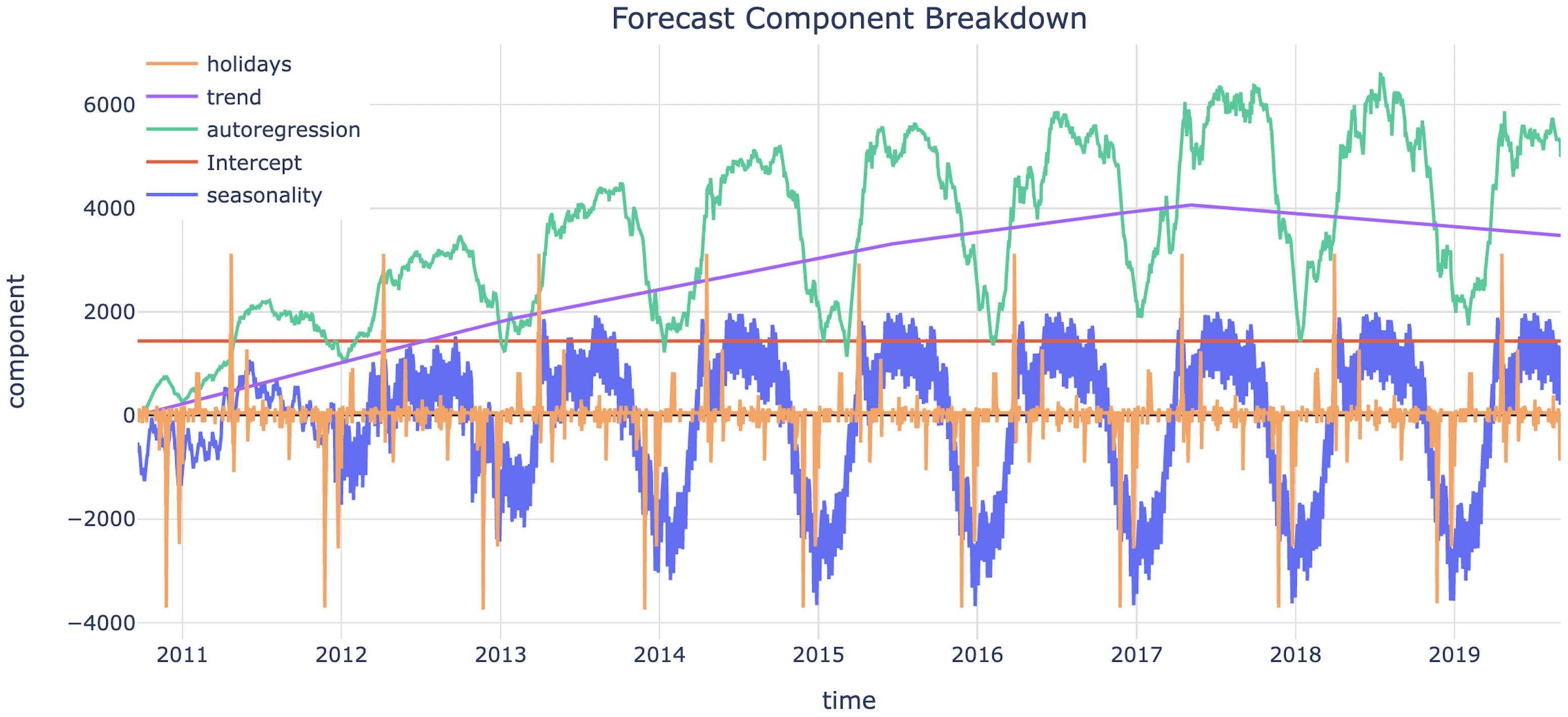}
	\vspace{-0.2in}
	\caption{The component plot shows how groups of covariates contribute to the forecasted value. It can be used for interpretability and debugging.
	}
	\label{fig:bikesharing_components}
	\Description{}
    \vspace{-0.1in}
\end{figure}

The last step is to diagnose the model and interpret results, both to improve the model
and to establish trust with stakeholders.
Again, we illustrate this on the D.C. bike-sharing dataset.
Figure~\ref{fig:bikesharing_components} plots forecast components such as
trend, seasonality, autoregression, and holidays, representing the forecast as a sum of the contribution from each group.
This view helps stakeholders understand the algorithm's assumptions (how it makes predictions)
and the patterns present in the dataset.
In Figure~\ref{fig:bikesharing_components}, the fitted trend first increases then slightly decreases after
a few detected changepoints.
The yearly seasonality reflects a higher number of rides during warmer months and a lower number of rides 
during colder months, with increasing magnitude over time.
In the presence of multicollinearity, one should treat this plot as descriptive of the model rather than
showing the true effect of each component. Effect interpretation is improved through
(1) groups of covariates that capture mostly orthogonal effects,
(2) regularization,
(3) fewer covariates when data are limited, and
(4) enough training data to distinguish effects.

\begin{figure}[h]
    \vspace{-0.05in}
    \centering
    \includegraphics[width=\linewidth]{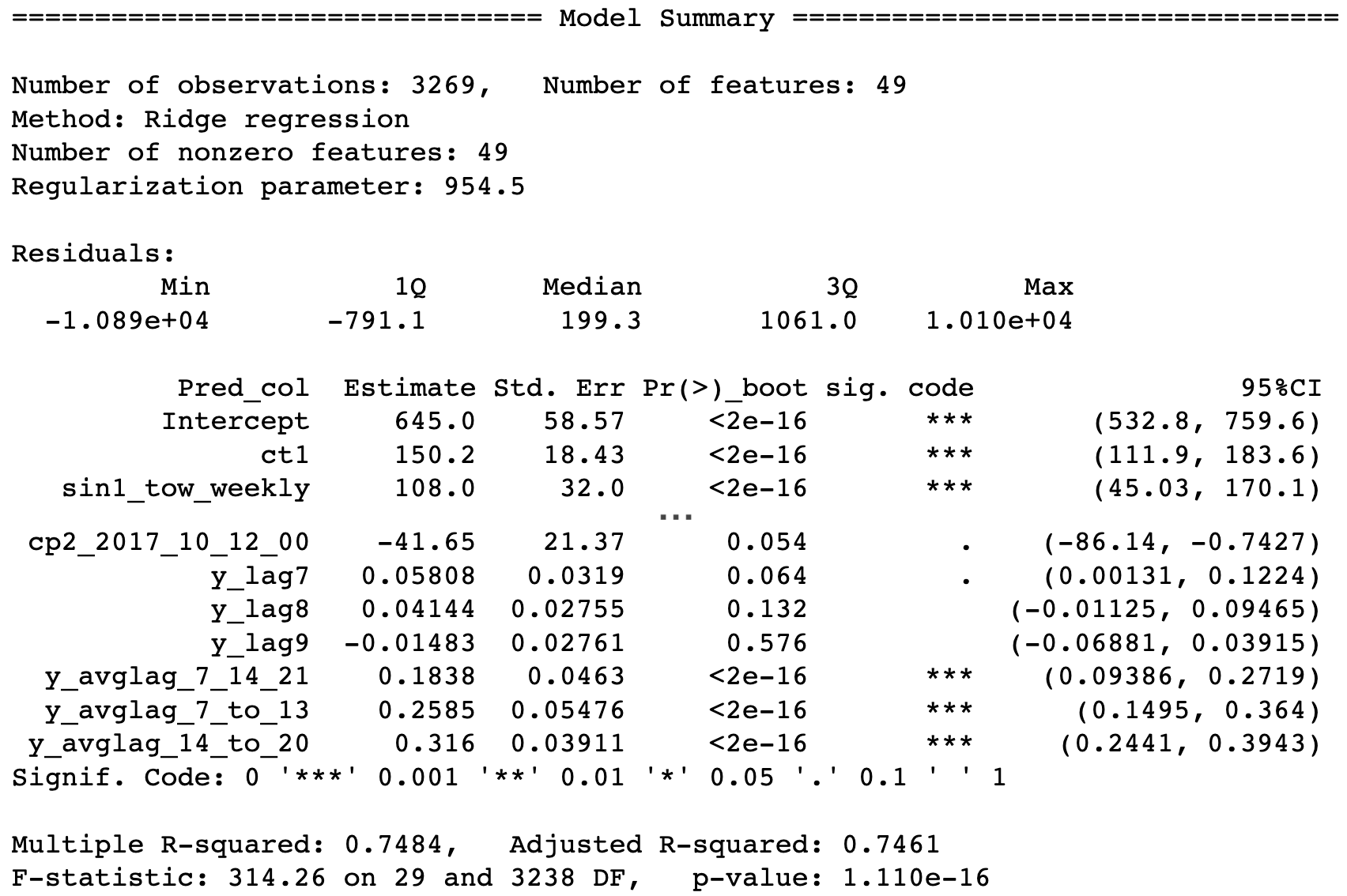}
     \vspace{-0.2in}
    \caption{Model summary shows the effect of individual covariates for interpretability, model tuning, and debugging.}
    \label{fig:bikesharing_summary}
    \Description{}
    \vspace{-0.1in}
\end{figure}

While component plot shows the effect of groups of covariates, model summary shows the
effect of individual covariates, as shown in Figure~\ref{fig:bikesharing_summary}.
It includes model overview, coefficients, p-values, and confidence intervals.
\texttt{Greykite} supports model summary for OLS and regularized models from \texttt{scikit-learn} and \texttt{statsmodels},
calculating the intervals using bootstrap for ridge and multi-sample splitting for lasso \cite{efron1994introduction, dezeure2015high}. This diagnostic can be used to inspect coefficients and assess whether any terms might be removed.

Component plot and model summary provide explanations of why forecasts change after
the model is trained on new data.
When strategic business decisions depend on the forecasted value, it is particularly important
to assess whether the drivers are reasonable before deciding to take action.

Thus, \texttt{Greykite}'s modeling framework addresses three key requirements of self-serve
forecasting adoption: (1) accuracy, (2) ease of use, (3) trust.
It achieves accuracy by making the Silverkite's flexible algorithms easy to configure and tune, ease of use by
aiding each step of the modeling process, and trust through interpretability.

    \section{Benchmark}
\label{sect:benchmark}
In this section, we compare Silverkite against Prophet and SARIMA (seasonal ARIMA) on a variety of datasets.
Prophet is a popular univariate forecasting algorithm designed for interpretability and
self-serve in the business context \cite{taylor-prophet-2018}.
SARIMA is a well-known forecasting model that has stood the test of time and is widely available
in many languages. While the model is not readily interpretable, it captures seasonality,
temporal-dependence, and non-stationarity and has been a strong baseline in recent
forecasting competitions \cite{makridakis2018m4}.
For the benchmark, we use the python packages
\texttt{greykite} v0.4.0,
\texttt{prophet} v1.0.1, and
\texttt{pmdarima} v1.8.0 \cite{pmdarima}. 

\subsection{Benchmark Setup}

Due to temporal dependency in time series data,
the standard $K$-fold cross-validation is not appropriate.
Rolling origin cross-validation is a common technique in the time series literature \cite{tashman2000out}.
Unlike ``fixed origin,'' which fits the data on the training set and forecasts on the following test period only once,
the rolling origin method evaluates a model with a sequence of forecasting origins that keeps moving forward.
This makes model evaluation more robust by averaging across time periods.
We use $K$-fold rolling origin evaluation, described in Figure~\ref{fig:benchmarking} in Appendix~\ref{sec:rolling_origin}.

To evaluate multiple time series at different scales, scale-independent metrics are frequently used,
such as MAPE, sMAPE \cite{flores1986pragmatic}, and Mean Absolute Scaled Error (MASE) \cite{hyndman2006another}.
The first two metrics have the disadvantage of being infinite or undefined when there are zero values in the data. 
In addition, they put a heavier penalty on errors for low actual values, hence they are biased.
Thus, we compare MASE with seasonal period according to the data frequency.
For calculation details, see Appendix~\ref{sec:mase_explained}.

\subsection{Experiments}

\subsubsection{Dataset and Setup}

Comparing time series forecasting models has gained interest in recent years,
such as in the M Competitions \cite{makridakis2018m4, makridakis2022m5}.
The curation of time series datasets has also grown rapidly,
including UCI Machine Learning Repository (with 100+ datasets) \cite{dua2019uci}, 
and the recently released Monash time series archive (30 datasets from different domains) \cite{godahewa2021monash}. 
\texttt{Prophet} and \texttt{Greykite} packages also have built-in datasets.

Many dataset collections contain either multivariate time series 
for global modeling or hundreds of univariate time series. 
They are intended for fixed origin evaluation,
and the result is averaged across many datasets.
In our benchmark framework, we intend to evaluate the algorithms over
a comprehensive period of time using a large number of splits.
Thus, we choose nine datasets from the above sources suitable for
rolling origin evaluation. The chosen datasets span a broad range of categories:
energy, web, economy, finance, transportation, and nature.
Their frequencies range from hourly to monthly
and we benchmark them across multiple forecast horizons. 
The datasets are summarized in Table~\ref{tab:dataset_setup}, with their detailed descriptions in Appendix~\ref{sec:benchmark_dataset}.
Any missing values are imputed by linear interpolation for model training.
The imputed values are not used when calculating model accuracy.

\begin{table}[h]
    \begin{tabular}{|l|l|l|l|l|}
         \hline
         (Frequency) Dataset & Length & Test Length & $K$ Splits \\
         \hline\hline
         (H) Solar power~\cite{godahewa2021monash} & 1 yr. & 30 days & 30*6 \\
         \hline
         (H) Wind power~\cite{godahewa2021monash} & 1 yr. & 30 days & 30*6 \\
         \hline
         (H) Electricity~\cite{dua2019uci} & 3 yr. & 365 days & 365*6 \\
         \hline
         (H) SF Bay Area traffic~\cite{godahewa2021monash} & 2 yr. & 60 days & 60*6 \\
         \hline
         (D) Peyton Manning~\cite{taylor-prophet-2018} & 9 yr. & 365 days & 365 \\
         \hline
         (D) Bike sharing~\cite{bikeshare} & 8 yr. & 365 days & 365 \\
         \hline
         (D) SF Bay Area traffic~\cite{godahewa2021monash} & 2 yr. & 365 days & 365 \\
         \hline
         (D) Bitcoin transactions~\cite{godahewa2021monash} & 11 yr. & 365 days & 365 \\
         \hline
         (M) Sunspot~\cite{godahewa2021monash} & 203 yr. & 24 months & 24 \\
         \hline
         (M) House supply~\cite{fred-housing} & 59 yr. & 24 months & 24 \\
         \hline
    \end{tabular}
    \caption{Datasets and their benchmark configuration. We use forecast horizons 1 and 24 for hourly data;
    1, 7, and 30 for daily data; 1 and 12 for monthly data.}
    \label{tab:dataset_setup}
    \vspace{-0.2in}
\end{table}

\vspace{-0.1in}
\subsubsection{Model Comparison}

We benchmark the performance of Silverkite, Prophet, and SARIMA.
For Silverkite, we use the ``AUTO'' model template.
For SARIMA, we use the out-of-the-box settings of \texttt{pmdarima}.
It uses statistical tests and AIC to identify the optimal $p, d, q, P, D, Q$ parameters. \footnote{\url{https://alkaline-ml.com/pmdarima/modules/generated/pmdarima.arima.AutoARIMA.html\#pmdarima.arima.AutoARIMA}}
For Prophet, we use the out-of-the-box settings with holidays added for hourly and daily frequencies to match the configuration of Silverkite.
Because Prophet also natively supports holidays, we use the same default holiday
countries for a fair comparison: US, GB, IN, FR, CN.
Holidays are not needed for monthly data due to the level of aggregation, so they are not included for monthly frequency.

\begin{table}[h]
    \vspace{-0.1in}
    \begin{tabular}{|l|l||l|l|l|}
         \hline
         \textit{MASE} & Horizon & Silverkite & Prophet & SARIMA \\
         \hline\hline
         {Hourly} & 1 & \textbf{0.741} & 2.047 & 1.178 \\
         \cline{2-5} & 24 & \textbf{0.945} & 2.034 & 3.429 \\
         \hline
         {Daily} & 1 & \textbf{0.940} & 1.527 & 1.358 \\
         \cline{2-5} & 7 & \textbf{1.097} & 1.539 & 1.515 \\
         \cline{2-5} & 30 & \textbf{1.251} & 1.575 & 1.705 \\
         \hline
         {Monthly} & 1 & 0.333 & 1.289 & \textbf{0.303} \\
         \cline{2-5} & 12 & \textbf{0.579} & 1.292 & 0.724 \\
         \hline
    \end{tabular}
    \caption{Model comparison. Average MASE across datasets. The best model for each frequency and horizon is in bold.}
    \label{tab:model_comparison_long}
    \vspace{-0.2in}
\end{table}

For each frequency and forecast horizon combination,
Table~\ref{tab:model_comparison_long} shows the MASE (lower is better)
averaged across benchmark datasets.
Silverkite significantly outperforms the other two algorithms in all
but one setting. For that setting (monthly frequency with horizon 1), Silverkite is
optimal on sunspot data and a close second on house supply data. 
The full results for each dataset are in Table~\ref{tab:full_result} in
Appendix~\ref{sec:full_results}.
Of the 24 dataset/horizon combinations, Silverkite is optimal on 20.
Because Silverkite has the advantage of being the most flexible of the three models,
it is possible to fine-tune Silverkite to achieve better performance on the
rest.

For a given frequency, Silverkite and SARIMA show noticable improvement on
shorter horizons, but Prophet does not. This suggests that Silverkite and SARIMA
make better use of recent information through autoregression,
whereas Prophet focuses on overall trends.
Silverkite's interpretable approach, with groups of covariates that capture underlying time series dynamics, could be one reason it outperforms SARIMA.

It is noteworthy that Prophet and SARIMA usually have MASE > 1 (i.e. forecast error is higher than the naive method's in-sample
error). Silverkite, on the other hand, usually has MASE < 1.
Thus, Silverkite offers good out-of-the-box performance on a wide range
of datasets, frequencies, and horizons.
Moreover, Silverkite is often faster than the other algorithms.
The runtime comparison for training and inference can be found in Tables~\ref{tab:runtime_train} and~\ref{tab:runtime_test} in Appendix~\ref{sec:runtime}.

    \section{Deployment}
\label{sect:deployment}
We deployed model-driven forecasting solutions at LinkedIn for more than twenty use cases
to streamline product performance management and optimize resource allocation.

\subsection{Performance Management}
The performance management process at Linkedin involves setting targets for core business metrics,
making investments, tracking progress, detecting anomalies,
and finding root causes.
Given its centrality and ubiquity across the business, it is critical to move away from
manual processes and adapt to growing data volume and complexity.
To achieve this, we facilitate adoption of \texttt{Greykite}.

First, we partnered with Financial Planning \& Analysis (FP\&A) to build automated forecasting-based performance
management for LinkedIn Marketing Solutions (LMS). Success for this customer was key to extending our outcomes
and learnings to more use cases for other business verticals.

LMS is a fast-growing business, with a complex ads ecosystem for advertisers and potential customers. Ads marketplace metrics exhibit complex growth and seasonality patterns and need to be modeled at different frequencies and across many dimensions. Due to this complexity, this use case became an essential learning experience for us to deploy new solutions and scale them to other business verticals. It helped us define requirements, for example:
\begin{itemize}
    \item Autoregression to improve next-day forecast accuracy.
    \item Indicators to capture sharp changes around month/quarter start and end.
    \item Trend and seasonality changepoints to capture the effect of large feature launches.
    \item User-provided anomalies to ignore temporary issues.
\end{itemize}
To enable efficient detection of revenue issues, our forecasts needed to meet a high level of accuracy (e.g.\;$<2\%$ MAPE for daily next-day forecast) and have reliable prediction intervals. An internal benchmark showed that Silverkite met the requirements and has $75\%$ lower error on revenue metrics compared to Prophet.

We deployed models for over thirty LMS metrics such as ads revenue and ads impressions and their key dimensions.
The forecasts are at daily, weekly, and monthly frequencies with horizons from 1 day to 6 months.
These are integrated into production dashboards and are sent in daily emails to show business outlook
and flag concerns about business health. The emails compare forecasted metrics against their targets and
alert anomalies in observed data compared to the forecasted baseline interval.
To aid investigation when revenue falls outside expectations, the dashboard includes forecasts for
supply-side and demand-side marketplace metrics and their dimensions. This helps isolate the problem
to a particular part of the supply or demand funnel or segment of the business.
The automated performance management solution has been in production for 18 months and is the primary source for
FP\&A to quickly assess business health and begin investigations if needed.

\begin{figure}[h]
    \vspace{-0.05in}
    \centering
    \includegraphics[width=\linewidth]{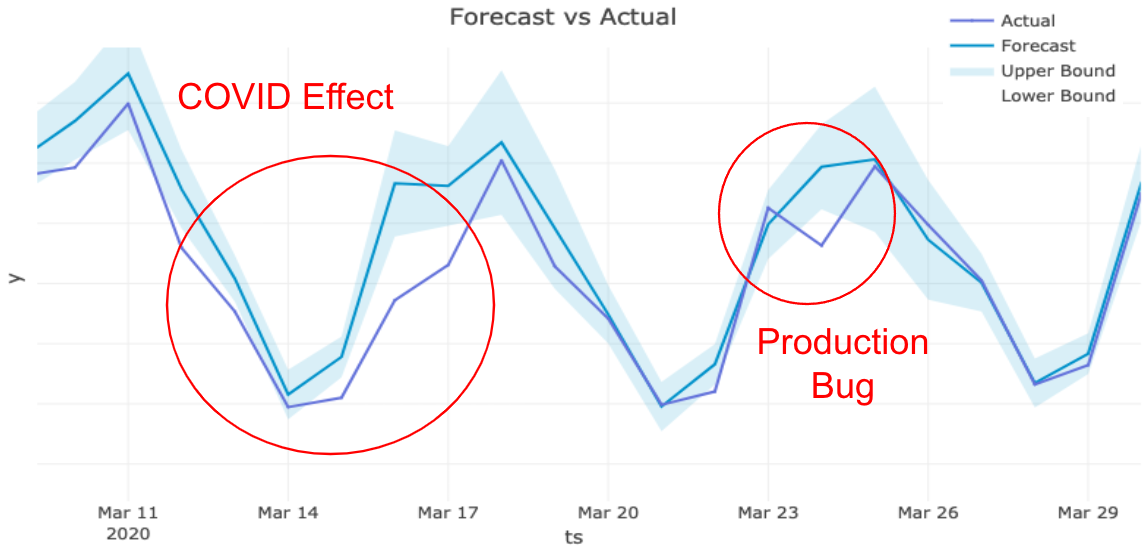}
    \vspace{-0.2in}
    \caption{\texttt{Greykite} detected unexpected anomalies and helped LinkedIn Premium assess impact severity.}
    \label{fig:anomaly_detection_POET_usecase}
    \Description{}
    \vspace{-0.08in}
\end{figure}

To scale our solution to other business verticals, we partnered with LinkedIn Premium and Talent Solutions to
provide short-term dimensional forecasts for key metrics such as sign-ups, bookings, sessions, and weekly
active users. We observed significant improvements in forecasting accuracy (MAPE) and faster anomaly detection with
higher precision and recall.
Figure~\ref{fig:anomaly_detection_POET_usecase} shows how \texttt{Greykite} detected anomalies for LinkedIn Premium and helped
them assess impact severity; these anomalies were missed by the existing week-over-week threshold detection.

\begin{figure}[h]
    \vspace{-0.1in}
    \centering
    \includegraphics[width=\linewidth]{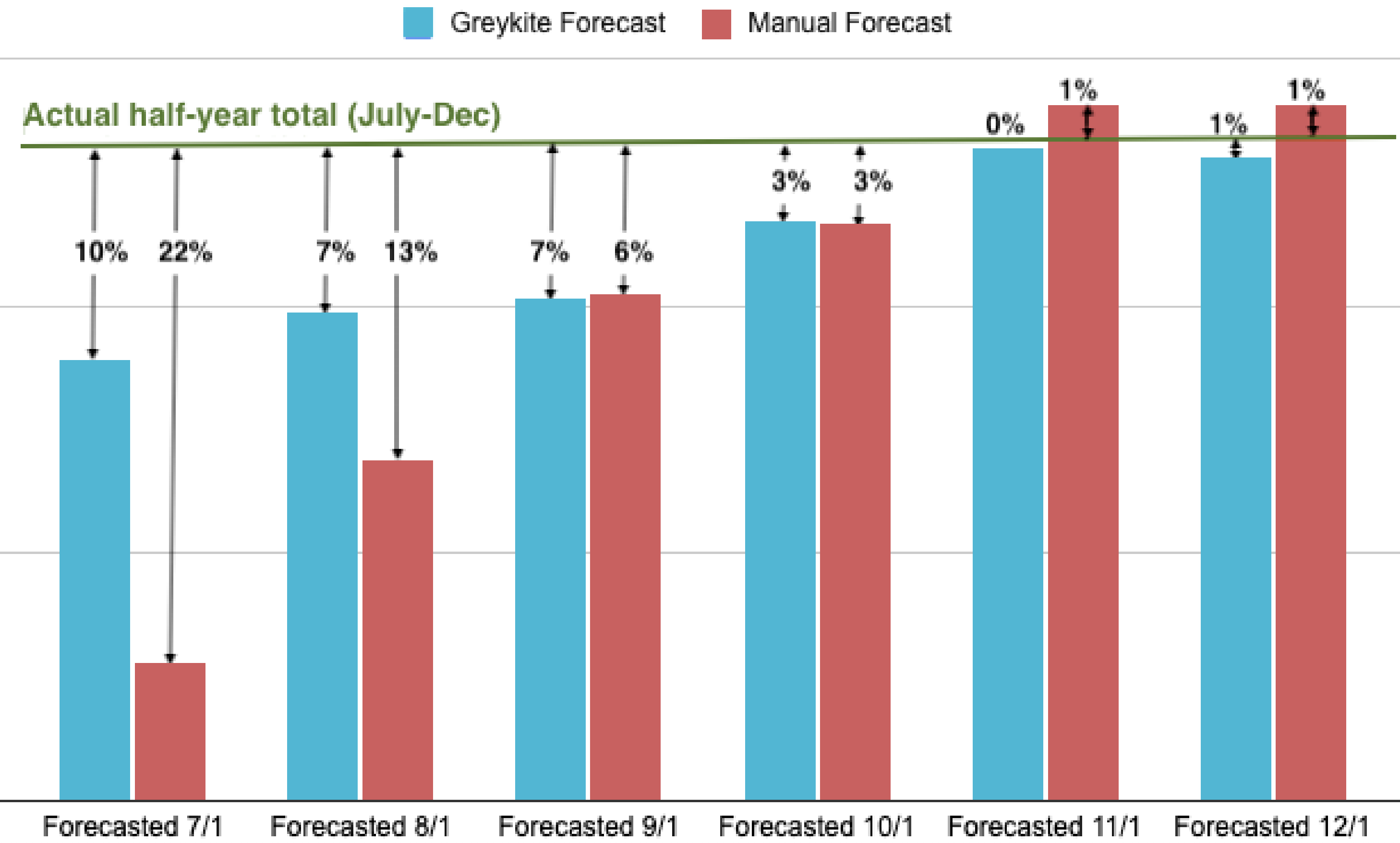}
    \vspace{-0.2in}
    \caption{\texttt{Greykite} outperforms manual forecast by picking up revenue growth momentum at the start of the half-year.}
    \label{fig:moneyball-outperform-manual-forecast}
    \Description{}
    \vspace{-0.15in}
\end{figure}

\texttt{Greykite} achieves remarkable performance on long-term forecasts as well. In partnership with FP\&A, we developed monthly forecasts for the next half-year of revenue that significantly outperformed manual forecasts, providing better guidance for strategic decisions.
Figure~\ref{fig:moneyball-outperform-manual-forecast} shows how the forecast adapted to revenue growth momentum earlier than the manual forecast through autoregression and automatic changepoint detection.

\subsection{Resource Management}
On the infrastructure side, forecasts help LinkedIn maintain site and service availability
in a cost-effective manner as site traffic continues to increase.
Better projections about future traffic, combined with accurate site
capacity measurements, enable confident decision-making and sustainable growth through
right-sized applications whose provisioned instances match required workloads.

\begin{figure}[h]
    \vspace{-0.1in}
    \centering
    \includegraphics[width=\linewidth]{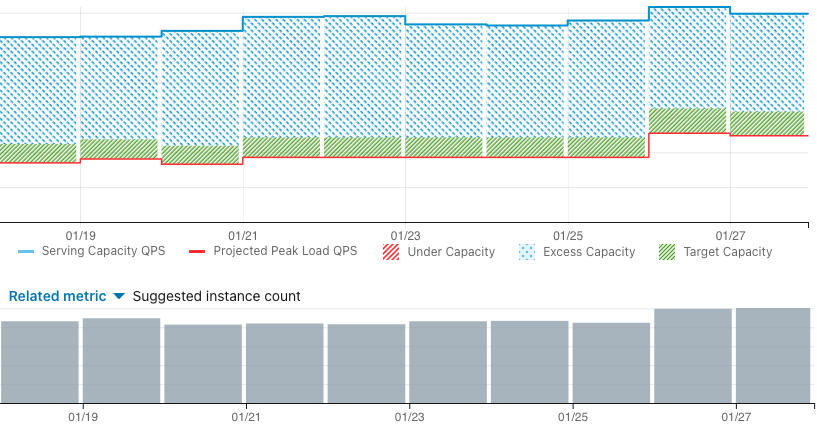}
    \vspace{-0.2in}
    \caption{The production dashboard for application headroom shows projected peak load QPS (red) and excess capacity (shaded blue) to inform decision-making on resource allocation. This plot reveals excess capacity that could be removed while safely supporting peak load.}
    \label{fig:headroom-dashboard}
    \Description{}
    \vspace{-0.2in}
\end{figure}

Prior to deploying \texttt{Greykite} forecasts, capacity planning was highly manual, reactive, and had a
tendency to overprovision resources. We provide hourly peak-minute queries per second
(QPS) forecasts for hundreds of services, which are used to automate right-sizing.
With the Automatic Rightsizing system conducting hundreds of rightsize actions a day,
we eliminate most of the toil required of application owners
to manually adjust compute resources to support organic business growth, save
millions of dollars by removing excess capacity, and optimize fleet resource utilization by
re-purposing such excess capacity to applications lacking capacity.
The forecasts are shown in production dashboards alongside allocated serving
capacity, as shown in Figure~\ref{fig:headroom-dashboard}.
The system has been in production for over two years. Since this collaboration,
we have seen significant reduction of capacity-related incidents
as applications are automatically uplifted to match expected workload.

\subsection{Lessons Learned}
We learned some lessons on algorithms, framework, and deployment through solving company-wide forecasting problems at scale.

\textbf{Diverse requirements can be met with a highly flexible family of models}. The flexible design that captures trend, seasonality, changepoints, holidays, autoregression, interactions, and regressors enables accurate forecasts of business and infrastructure metrics across frequencies and horizons.

\textbf{It is possible to achieve high accuracy without sacrificing interpretability}. Interpretation is essential to building trust with stakeholders who want to understand how forecasts are made, modify assumptions, and understand why forecasted values change after training on more data.
Silverkite transforms features into a space that can be used in a regularized linear model, allowing additive component plots and model summary.

\textbf{Enabling self-serve allowed scaling forecast solutions across the business}. This required high out-of-the-box accuracy with little effort, which we achieved with an end-to-end forecasting library and intuitive tuning interfaces, and a fast algorithm for interactive tuning and hyperparameter search.

\textbf{Partnership with customers is essential to deploying new research ideas}. During alpha and beta, we worked with a few champion customers to clarify requirements and deliver wins. Then, we generalized the solution and scaled it through a self-serve model. Champion customers should be high impact, representative of other customers, aligned with company priorities, and willing to form a close partnership. As we proved success, the champion use cases became advocates and examples to drive adoption forward.

    \section{Discussion}
\label{sect:discussion}
Prior to this work, LinkedIn's forecasts were mostly manual, ad-hoc, and intuition-driven.
Now, customers from all lines of business and all functions (engineering, product, 
finance) are adopting algorithmic forecasts.
Customers understand the benefits of accuracy, scale, and consistency. Our forecasts save time and drive clarity on the business/infrastructure that empowers LinkedIn to plan and adapt
dynamically to new information.
This culture shift was achieved through a fast, flexible, and interpretable family of models and
a modeling framework that enables self-serve forecasting with ease and accuracy.

The Silverkite algorithm performs well on both internal datasets and out-of-the-box on external datasets
with frequencies from hourly to monthly from a variety of domains.
Its flexible design allows fine-tuning of covariates, objective
function, and volatility model.
Thus, we expect the open-source \texttt{Greykite} library to be useful to forecast practitioners,
especially those whose time series include time-varying growth and seasonality,
holiday effects, anomalies, and/or dependency on external factors,
characteristics that are common in time series related to human activity.

\begin{acks}
    This work is done by the Data Science Applied Research team at LinkedIn.
    Special thanks to our TPM Shruti Sharma and our collaborators in
    Data Science, Engineering, SRE, FP\&A, BizOps, and Product for adopting the library.
    In particular, thanks to
    Ashok Sridhar, Mingyuan Zhong, Peter Huang,
    Hamin Oh, Neha Gupta, Neelima Rathi,
    Deepti Rai, Christian Rhally, Camilo Rivera, Priscilla Tam,
    Meenakshi Adaikalavan, Zheng Shao,
    Mike Snow, Stephen Bisordi, Dong Wang, Ankit Upadhyaya,
    and Rachit Kumar
    for allowing us to share their use cases.
    We also thank our leadership team Ya Xu and Sofus Macsk\'{a}ssy for their support.
\end{acks}

\bibliographystyle{ACM-Reference-Format}
\bibliography{greykite_bibliography.bib}

\clearpage

\appendix
\section{Appendix}
\label{sect:appendix}

\subsection{Rolling Origin Evaluation}
\label{sec:rolling_origin}
\texttt{Greykite} includes the \texttt{BenchmarkForecastConfig} class to perform rolling origin benchmarking.
Rolling origin evaluation works as follows.
We replicate the time series data to $K$ train-test splits.
Each split contains a test period (red) immediately after its training period (deep and light blue).
The $K$ forecasting origins (test start dates) are picked in a rolling manner,
working backward from the end of the time series.

Each candidate model is fit on the training period of each of the rolling splits,
and the trained model is used to predict on the corresponding test period. 
The average test error across all $K$ splits mimics the backtest error if
the model were put in production, and this provides a reliable estimate of
the future forecast error.

\begin{figure}[h]
    \vspace{-0.1in}
    \centering
    \includegraphics[width=\linewidth]{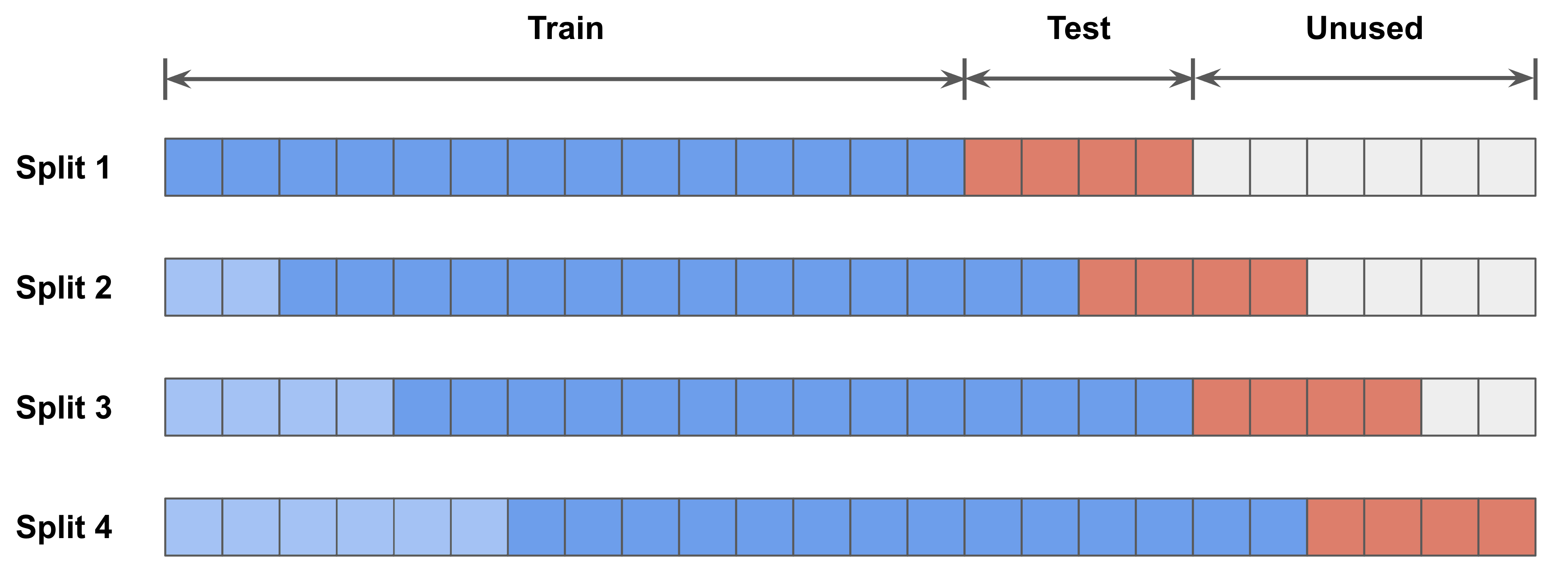}
    \caption{$K$-fold rolling origin evaluation. The example shows four splits with forecast horizon 4 and period between splits 2.
    The light blue points could either be included or not included in the training set, depending on user's choice of whether to use expanding windows (light blue + deep blue) or moving windows (deep blue only).}
    \label{fig:benchmarking}
    \vspace{-0.1in}
\end{figure}

A few parameters control how the rolling origin splits are generated. 
First, since one of the goals of rolling origin evaluation is to make sure
the estimated forecast error is robust and representative,
$K$ needs to be sufficiently large.
We recommend that the test periods cover at least one year of data for daily and less granular frequencies (weekly, monthly, etc.).
For sub-daily data with a short forecast horizon, we recommend the test periods cover at least one month.

$K$ should be chosen together with the period between the splits, which controls the step size between every two successive splits.
In general, we recommend step size of $1$ to obtain the most comprehensive evaluation. 
In practice, for small granularity data (such as hourly or minutely), 
it can make sense to increase the period between splits and reduce $K$
to decrease computational cost.
As a result, the test periods in these rolling splits may or may not overlap.

Another two parameters are the lengths of the training and test periods
for each split. In order to best estimate the model's performance in real applications,
the length of each test period should match the desired forecast horizon.
Ideally, the minimum training period is at least two years for daily
and less granular frequencies in order to accurately estimate the yearly seasonality effect.
For sub-daily data, a shorter training period could be used for short-term forecasts.
The model that minimizes the average error across all the rolling splits is deemed as the best model.

Finally, the split training periods could either use expanding windows (fixed train start date)
or moving windows (fixed length), as shown in Figure~\ref{fig:benchmarking}.
We recommend choosing the one that best mimics the setting for deployment.
Moving windows is preferred to increase speed when not all history is needed for an accurate forecast.
Expanding windows is preferred if speed is not an issue and using more history improves accuracy.

Such an extensive rolling evaluation not only provides a robust estimate of the model performance, 
but also enables users to run comprehensive diagnostics, 
such as error breakdowns on different seasonal cycles (e.g. day of week) and holiday effects.
Then, these systematic errors could be addressed by incorporating these signals as features into the model.
We developed an internal product \texttt{Greykite-on-Spark} for benchmarking and hyperparameter tuning on Spark.
With parallel execution on Spark, all experiments could be finished within a day.

\subsection{Evaluation Metrics}
\label{sec:mase_explained}
MASE \cite{hyndman2006another} is calculated by dividing the out-of-sample MAE by the in-sample MAE from a naive forecast.
For daily time series with weekly seasonality, the naive forecast uses the value at $T-7$ to predict the value at time $T$,
where 7 is called the “seasonal period”. For non-seasonal time series, the seasonal period is 1.
Similar to \cite{godahewa2021monash}, we choose the seasonal period based on the frequency of the data.
In our experiments, we use 24 for hourly data, 7 for daily data, and 12 for monthly data.

\subsection{Benchmark Datasets}
\label{sec:benchmark_dataset}

\textbf{Hourly}. 4 datasets, horizons 1 and 24.
The hourly datasets are solar power, wind power,
electricity, and traffic. These datasets
are from Monash~\cite{godahewa2021monash}.

The solar (wind) power dataset contains the solar (wind) power production of 
an Australian wind farm from August 2019 to July 2020, with original frequency 4-second. 
We aggregate it to an hourly series and remove any incomplete hours. Since the time series is shorter than one year,
the seasonal effects may not be well estimated. 
Also, the solar power dataset is an intermittent time series
where a lot of zeros are present, hence using MASE is better than MAPE or sMAPE.
We use a moving 300-day window for training, and test on every 4 hours in the last 30 days (30*6 splits).

The electricity dataset contains the hourly consumption (in kW) of 321 clients from 2012 to 2014
published by Monash \cite{godahewa2021monash}, originally from \cite{dua2019uci}. 
We aggregate them by taking the average across the 321 clients. 
The averaged series has a history of 3 years. 
We use a moving two-year window for training, and test on every 4 hours in the last year (365*6 splits).

The SF Bay Area traffic dataset records the road occupancy rates (between 0 and 1) measured by different
sensors on San Francisco Bay area freeways from 2015 to 2016. We take the average of these occupancy rates.
We use a moving 600-day window for training, and test on every 4 hours in the last 60 days (60*6 splits).

These datasets present both intraday and interday seasonalities. 

\textbf{Daily}. 4 datasets, horizons 1, 7, and 30.
The daily datasets are page views, bike rental counts, traffic,
and Bitcoin transactions.
We test on the last year (365 splits, period between splits 1), with an expanding window for training.

The Peyton Manning dataset contains the logarithm of daily page views 
for the Wikipedia page for football player Peyton Manning. 
This dataset is from \texttt{Prophet} \cite{taylor-prophet-2018} and it has 8 years of history.
Missing values are imputed by us with linear interpolation when fitting the models.

The bike-sharing dataset contains aggregated daily counts of the number of rented bikes
in Washington, D.C. from 2010 to 2019 (incomplete days are removed).
The raw dataset is from Capital Bikeshare \cite{bikeshare}
and it is also available in \texttt{Greykite}.

The SF Bay Area traffic dataset is the same as for hourly benchmarking.
We aggregate it to a daily time series from 2015 to 2016.

The Bitcoin dataset has the number of Bitcoin transactions from 2009 to 2021.
The dataset was curated (with missing values filled)
by Monash \cite{godahewa2021monash}.

The Peyton Manning and bike-sharing datasets present weekly and yearly seasonalities and possible changepoints.
The SF Bay Area traffic dataset also shows relatively strong weekly and yearly seasonalities but has a shorter history,
which makes it harder to estimate long-term effects.
The Bitcoin dataset is much more volatile than the others.

\textbf{Monthly}. 2 datasets, horizons 1 and 12.
The monthly datasets are sunspot activity and house supply.
We test on the last 2 years of data (24 splits, 1 period between splits), with an expanding window for training.

The first dataset is daily sunspot activity from 1818 to 2020 published by
Monash \cite{godahewa2021monash}, originally from \cite{clette2014revisiting}.
We aggregate it to a monthly time series more than 200 years long.
The second dataset is monthly house supply from 1963 to 2021 obtained from FRED \cite{fred-housing}.
Both have mild yearly seasonality and some multi-year patterns.

\subsection{Full Benchmark Results}
\label{sec:full_results}

The complete benchmark results (MASE) are shown in Table~\ref{tab:full_result}.

\begin{table}[h]
    \begin{tabular}{|l|l||l|l|l|}
        \hline
        \textit{MASE} & Horizon & Silverkite & Prophet & SARIMA \\
        \hline\hline
        {(H) Solar power} & 1 & \textbf{0.732} & 1.760 & 1.397 \\
        \cline{2-5} & 24 & \textbf{0.854} & 1.721 & 3.794 \\
        \hline
        {(H) Wind power} & 1 & 0.220 & 0.569 & \textbf{0.211} \\
        \cline{2-5} & 24 & 0.742 & 0.706 & \textbf{0.626} \\
        \hline
        {(H) Electricity} & 1 & \textbf{0.820} & 3.168 & 1.323 \\
        \cline{2-5} & 24 & \textbf{1.030} & 3.403 & 6.738 \\
        \hline
        {(H) SF Bay Area} & 1 & \textbf{1.191} & 2.689 & 1.779 \\
        \cline{2-5} traffic & 24 & \textbf{1.153} & 2.306 & 2.558 \\
        \hline
        {(D) Peyton} & 1 & \textbf{0.646} & 0.826 & 0.709 \\
        \cline{2-5} Manning & 7 & \textbf{0.810} & 0.827 & 0.884 \\
        \cline{2-5} & 30 & \textbf{0.792} & 0.842 & 1.116 \\
        \hline
        {(D) Bike sharing} & 1 & \textbf{0.843} & 1.002 & 0.927 \\
        \cline{2-5} & 7 & \textbf{0.964} & 1.010 & 1.031 \\
        \cline{2-5} & 30 & \textbf{1.006} & 1.029 & 1.166 \\
        \hline
        {(D) SF Bay Area} & 1 & \textbf{0.712} & 0.840 & 1.921 \\
        \cline{2-5} traffic & 7 & \textbf{0.798} & 0.847 & 2.244 \\
        \cline{2-5} & 30 & \textbf{0.814} & 0.874 & 2.354 \\
        \hline
        {(D) Bitcoin} & 1 & \textbf{1.557} & 3.438 & 1.876 \\
        \cline{2-5} transactions & 7 & \textbf{1.816} & 3.470 & 1.902 \\
        \cline{2-5} & 30 & 2.392 & 3.554 & \textbf{2.184} \\
        \hline
        {(M) Sunspot} & 1 & \textbf{0.088} & 1.497 & 0.114 \\
        \cline{2-5} & 12 & \textbf{0.219} & 1.572 & 0.397 \\
        \hline
        {(M) House supply} & 1 & 0.577 & 1.080 & \textbf{0.492} \\
        \cline{2-5} & 12 & \textbf{0.939} & 1.012 & 1.050 \\
        \hline
    \end{tabular}
    \caption{Detailed model comparison. Showing MASE for each dataset and horizon. The best model for each row is in bold.}
    \label{tab:full_result}
\end{table}

\subsection{Runtime Comparison}
\label{sec:runtime}

\begin{table}[h]
    \begin{tabular}{|l|l||l|l|l|}
        \hline
        \textit{Runtime} & Train Length & Silverkite & Prophet & SARIMA \\
        \hline\hline
        Hourly & 17520 & \textbf{30.57} & 142.39 & 173.86 \\
        \hline
        Daily & 2963 & \textbf{4.75} & 16.93 & 7.98 \\
        \hline
        Monthly & 2428 & 3.41 & \textbf{0.39} & 5.46\\
        \hline
    \end{tabular}
    \caption{Training time comparison (in seconds).}
    \label{tab:runtime_train}
\end{table}

\begin{table}[h]
    \begin{tabular}{|l|l||l|l|l|}
        \hline
        \textit{Runtime} & Horizon & Silverkite & Prophet & SARIMA \\
        \hline\hline
        Hourly & 1 & 0.88 & 1.39 & \textbf{0.00} \\
        \hline
        Daily & 1 & 0.51 & 2.08 & \textbf{0.00} \\
        \hline
        Monthly & 1 & 0.12 & 1.05 & \textbf{0.00} \\
        \hline
    \end{tabular}
    \caption{Inference time comparison (in seconds).}
    \label{tab:runtime_test}
\end{table}

We compare the training and inference runtime on a MacBook Pro with 2.4 GHz 8-Core Intel Core i9 processor, and 32 GB 2667 MHz DDR4 memory.
Results are recorded in Tables~\ref{tab:runtime_train} and~\ref{tab:runtime_test}.
Hourly models are evaluated with 2 years of training data using the
electricity dataset. Daily models are evaluated with 8 years of training data using the Peyton Manning dataset. Monthly
models are evaluated with 202 years of training data using the sunspot dataset. All models use forecast horizon 1.
All models use the same configurations as before to produce a forecast and 95\% prediction interval. All measurements are the best of 7 runs.

Silverkite's training speed is significantly faster than Prophet and SARIMA.
The only exception is the monthly Prophet model, which is extremely fast because its features for monthly data are quite limited.
However, Prophet's MASE for sunspot monthly data with horizon 1 is 17x higher than Silverkite's (1.497 vs 0.088).

Silverkite's inference speed is significantly faster than Prophet's for all frequencies because Prophet requires MCMC to sample from a distribution.
The inference time of SARIMA is near zero because its predictions are
easy to calculate from the analytical solution, whereas
Silverkite prepares the future features and performs a larger matrix multiplication.

\end{document}